\pdfoutput=1

\documentclass[11pt]{article}
\usepackage{hyperref}
\usepackage{comment}
\usepackage{multirow}
\usepackage{dingbat}
\usepackage[preprint]{acl}

\usepackage{times}
\usepackage{latexsym}

\usepackage{xurl}

\usepackage{booktabs}
\usepackage{multirow}

\usepackage[T1]{fontenc}

\usepackage[utf8]{inputenc}

\usepackage{microtype}

\usepackage{inconsolata}

\usepackage{graphicx}
\usepackage[most]{tcolorbox}
\usepackage{booktabs}
\usepackage{enumitem}
\setlist{nolistsep}
\setlist[itemize]{leftmargin=*}
\setlist[enumerate]{leftmargin=*}
\title{Financial Risk Relation Identification through Dual-view Adaptation}

\author{
 \textbf{Wei-Ning Chiu\textsuperscript{1,2}},
 \textbf{Yu-Hsiang Wang\textsuperscript{2}},
 \textbf{Andy Hsiao\textsuperscript{2}},
 \textbf{Yu-Shiang Huang\textsuperscript{1,2}},
\\
 \textbf{Chuan-Ju Wang\textsuperscript{2}}
\\
\\
 \textsuperscript{1}National Taiwan University,
 \textsuperscript{2}Academia Sinica
\\
 \small{
   \textbf{Correspondence:} \href{mailto:cjwang@citi.sinica.edu.tw}{cjwang@citi.sinica.edu.tw}
 }
}

\begin{document}
\pagestyle{empty}
\maketitle
\begin{abstract}
A multitude of interconnected risk events---ranging from regulatory changes to geopolitical tensions---can trigger ripple effects across firms.
Identifying inter-firm risk relations is thus crucial for applications like portfolio management and investment strategy.
Traditionally, such assessments rely on expert judgment and manual analysis, which are, however, subjective, labor-intensive, and difficult to scale. 
To address this, we propose a systematic method for extracting inter-firm risk relations using Form 10-K filings---authoritative, standardized financial documents---as our data source.
Leveraging recent advances in natural language processing, our approach captures implicit and abstract risk connections through unsupervised fine-tuning based on chronological and lexical patterns in the filings.
This enables the development of a domain-specific financial encoder with a deeper contextual understanding and introduces a quantitative risk relation score for transparency, interpretable analysis.
Extensive experiments demonstrate that our method outperforms strong baselines across multiple evaluation settings. Our codes are available at \url{https://github.com/cnclabs/codes.fin.relation}.
\end{abstract}

\section{Introduction}

Relation identification between entities is valuable across various domains---including healthcare, legal analytics, social networks---and finance is no exception.
The financial market is a complex ecosystem shaped by a wide array of factors, including economic indicators, geopolitical events, corporate developments, regulatory changes, and investor sentiment. 

Among various types of inter-entity links, risk relations are particularly important due to their implications for financial performance and decision-making. A risk relation exists when two companies are both exposed to the same risk factors—such as new regulations, ongoing lawsuits, supply chain disruptions, or broader economic downturns. In such cases, an adverse event affecting one firm can also influence the other, creating a shared vulnerability.
For example, both Nvidia (NVDA) and Wabtec (WAB) faced disruptions from the semiconductor supply chain crisis in 2022.\footnote{Nvidia: \url{https://www.reuters.com/technology/graphic-chip-price-drop-raises-questions-whether-end-shortage-is-sight-2022-04-25/}. \newline Wabtec: \url{https://www.nasdaq.com/articles/whats-in-the-offing-for-wabtec-wab-this-earnings-season}.}
Identifying such relations is vital for informed investment decisions.
However, traditional expert-driven analysis is often
time-consuming, subjective, and prone to cognitive biases such as overconfidence or herd behavior.

This creates a growing need for objective, scalable methods to extract risk-related connections from financial texts.
Structured documents like Form 10-K filings offer a rich, standardized resource for such analysis.\footnote{Form 10-K filings are annual reports mandated by the U.S. Securities and Exchange Commission (SEC).}
Recent advances in natural language processing (NLP), particularly pretrained language models, offer powerful tools for learning semantic representations.
Yet most existing relation extraction methods focus on explicit entity-relation tagging and struggle to capture the implicit or abstract connections---like shared risk exposures---prevalent in financial texts.

To overcome these limitations, we adopt a retrieval-based encoding framework---a foundation of modern NLP---that transforms text into dense, semantically rich vector representations.
This enables efficient relation discovery, information retrieval, and downstream financial NLP applications~\cite{wang2024bertir, alaparthi2020bert-sa}.
Specifically, we first propose an unsupervised fine-tuning strategy based on a dual-view similarity framework to adapt general-purpose encoders for the finance domain.
Our approach leverages two key characteristics of Form 10-K filings: (1) standardized language and constrained vocabulary, which yield consistent lexical patterns; and (2) frequent association of risk events with date-time references, enabling temporal alignment.
By modeling both lexical and chronological similarities, we construct high-quality positive training pairs that reflect semantically or temporally aligned content.
This dual-view supervision guides the encoder to capture nuanced financial semantics and align similar risk disclosures across firms.

\begin{figure*}[t]
  \centering
  \includegraphics[width=\textwidth]{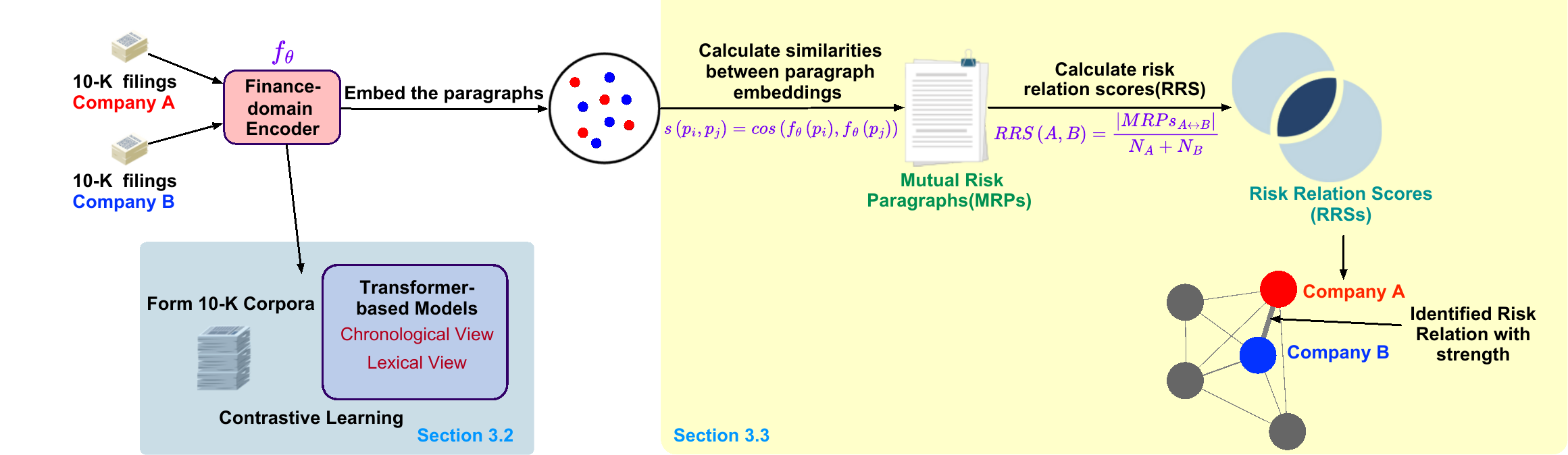}
  \caption{\bf {Overview of risk identification pipeline.}}
  \label{fig:overview}
\end{figure*}

To complement this encoder, we secondly introduce a retrieval-based, interpretable scoring mechanism---{\it the risk relation score} (RRS)---to quantify inter-firm risk connections. 
RRS offers key advantages over traditional heuristics: it is symmetric, guarantees minimum similarity thresholds, and enhances interpretability by grounding each relation in explicitly retrieved mutual risk paragraphs (MRPs).
This not only provides a robust measure of shared risk but also delivers textual evidence, improving the transparency and reliability of the model's output. Figure \ref{fig:overview} gives an overview of our method for identifying inter-firm risk relations.\footnote{A peer-reviewed system demonstration related to this work was independently developed and published as a demo paper at NAACL 2025 by~\citet{wang-etal-2025-surf}. Although both the demo paper and the present paper employ similar methodology, the former primarily emphasizes the visualization of outcomes, whereas the latter focuses on the methodological contributions and comprehensive empirical experiments across various aspects.}

Despite recent progress, evaluating inter-firm relations remains challenging due to the lack of standardized benchmarks and domain-specific evaluation protocols. 
To address this, we conduct comprehensive experiments to assess our method:
\begin{enumerate}
\item
We show that RRS correlates strongly with the absolute values of daily stock return correlations, demonstrating real-world relevance. 
\item 
We integrate the discovered risk relations into a graph-based model for stock price prediction, significantly improving its performance. 
\item
We assess the standalone retrieval capabilities of our encoder using \texttt{MultiHiertt}~\cite{zhao-etal-2022-multihiertt}, a financial QA benchmark built from regulatory filings, where our model consistently outperforms strong baselines.
\end{enumerate}

\paragraph{Contributions}
\begin{itemize}
\item \textbf{Risk Relation Scoring.} We introduce a novel metic---\emph{risk relation score (RRS)}---to measure inter-firm risk relationships.
Based on encoder-derived paragraph similarity, RRS is transparent, interpretable, and symmetric, with explicit textual evidence.
\item \textbf{Domain-specific Encoder Fine-tuning.} We propose a dual-view unsupervised fine-tuning strategy using lexical and chronological similarity patterns in Form 10-K filings to adapt general NLP encoders to the financial domain.
\item \textbf{Comprehensive Empirical Validation.} We demonstrate the utility of our approach through stock return correlation analysis, graph-based forecasting improvements, and strong retrieval performance on the \texttt{MultiHiertt} benchmark.
\end{itemize}

\section{Related Works}\label{sec:related}
In the field of natural language processing (NLP), pretrained encoders have been instrumental in transforming text into semantically rich representations.
This section reviews advances in relation extraction, general-domain encocers, and financial-domain encoders, with an emphasis on their relevance to financial text analysis.

\subsection{Relation Extraction}
Early relation extraction (RE) methods relied on pattern matching  and manual feature engineering, such as 
such as the lexico-syntactic patterns introduced by \citet{hearst1992hyponyms}.
With the rise of deep learning, neural models have like CNN~\cite{zeng-etal-2015-pcnn} and PCNN~\cite{zeng-etal-2015-pcnn} improved RE through automated feature learning. 
The advent of pretrained language models, particularly BERT~\cite{devlin-etal-2019-bert}, 
significantly advanced RE by enabling fine-tuning for contextual understanding. 
Approaches such as entity-aware fine-tuning~\cite{baldini-soares-etal-2019-mtb}  further boosted performance.
Recent surveys~\cite{diaz2024re-survey} highlighted the dominance of BERT-based techinques while recognizing the growing impact of large language models (LLMs) like T5~\cite{raffel2020t5}.

\subsection{General-domain Encoders}
Transformer-based models, led by BERT~\cite{devlin-etal-2019-bert} and its variants such as~RoBERTa~\cite{liu-etal-2019-roberta} and SpanBERT~\cite{joshi-etal-2020-spanbert}, transformed NLP by enabling transfer learning across tasks.
In the context of retrieval, DPR~\cite{karpukhin-etal-2020-dpr} introduced supervised dense retrieval with dual encoders, while Spider~\cite{ram2021spider} used instruction-tuned data to enhance performance. Contriever~\cite{izacard2021unsupervised} offered an unsupervised contrastive learning alternative, effectively capturing semantic similarity. 
More recent models, such as Jina AI's encoder~\cite{gunther2023jina}, incorporates novel methods like Attention with Linear Biases (ALiBi)~\cite{press2021alibi} to effectively process longer textual sequences. Embeddings in~\citet{xiao2024cpack} further refine encoding strategies by considering context from multiple perspectives including functionality, granularity, and linguity.

\subsection{Financial-domain Encoders}
Domain-specific encoders have been introduced to better capture the nuances of financial language.
FinBERT~\cite{araci2019finbert} adapted BERT by pretraining on large-scale financial corpora and has demonstrated superior performance in sentiment analysis.
SEC-BERT~\cite{loukas2022finer} further narrowed the focus by training exclusively on U.S. SEC filings, enhancing its applicability to regulatory documents.
Another variant, FinBERT-MRC~\cite{zhang2023finbert-mrc} reformulated financial named entity recognition as a machine reading comprehension task, improving contextual precision.
Beyond BERT-based models, Fin-E5~\cite{tang2025finmteb} introduced a persona-driven synthetic data strategy to support a wider range of financial embedding tasks.
In parallel, proprietary models such as BloombergGPT~\cite{wu2023bloomberggpt} showcased the potential of financial LLMs trained on exclusive datasets, though their closed nature spurred demand for open alternatives.
In response, FinGPT~\cite{yang2023fingpt} offers an open-source framework focused on accessible data and democratized financial LLMs.

\section{Methodology}
\label{sec:Methodology}
This section details our overall approach for identifying and quantifying inter-firm risk relations from financial documents.
We begin by introducing the notations and terminology used throughout the paper in Section~\ref{sec:notation}. 
We then present two core components of our methodology: (1) an unsupervised fine-tuning strategy that adapts a general encoder to better capture financial semantics (Section~\ref{sec:finetuning}, and (2) a risk relation scoring mechanism that leverages the fine-tuned encoder to compute transparent and symmetric measures of inter-firm risk exposure (Section~\ref{sec:scoring}).
Figure~\ref{fig:overview} illustrates the overall framework for identifying inter-firm risk relations.

\subsection{Notation and Preliminaries}\label{sec:notation}
Key notations used in this paper are defined below:
\begin{itemize}
\item $w$: A token, i.e., a word or subword unit, within a paragraph.
\item $p$: A paragraph, represented as a sequence of tokens: $p = [w_1, \dots, w_n]$, where $n$ is the number of tokens.
\item $D$: A batch of paragraphs.
\item $f_\theta$: An encoder function parameterized by $\theta$, which maps an input sequence to a dense vector representation.
\item $f_\theta(p)$: The vector representation of paragraph $p$, obtained by mean pooling over the final-layer hidden states of the encoder.
\item $s(p_i,p_j)$: The similarity score between paragraphs $p_i$ and $p_j$, typically computed using cosine similarity between their embeddings.
\item $\mathcal{P}_A$, $\mathcal{P}_B$: The sets of paragraphs associated with firms $A$ and $B$, respectively.
\end{itemize}

\begin{figure*}[t]
  \centering
  \includegraphics[width=0.85\textwidth]{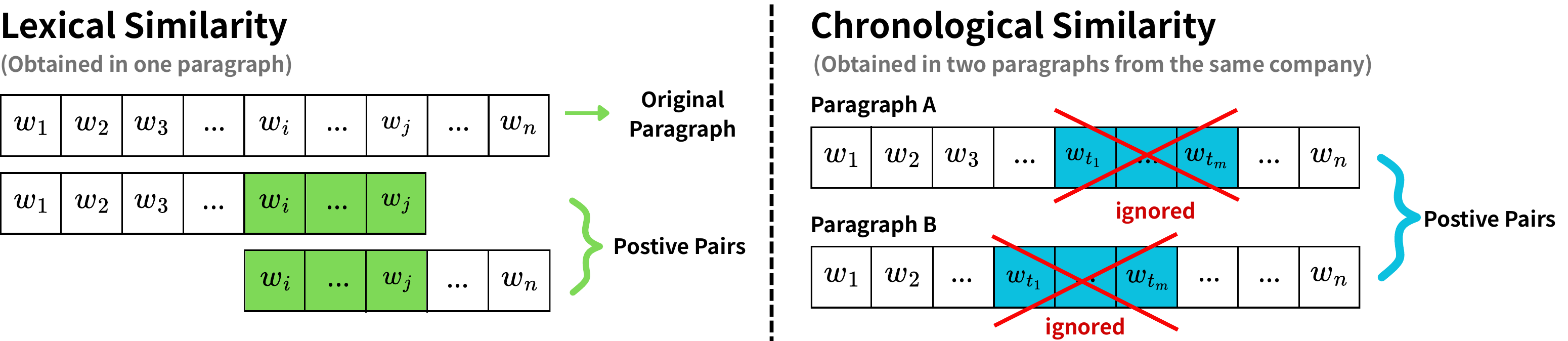}
  \caption{{\bf Illustration of the dual-view training strategy.} Left (lexical view): Two overlapping spans within the same paragraph are sampled (green) and treated as a positive pair. Right (chronological view): Two paragraphs from the same company that share identical date-time tokens form a positive pair; date‑time tokens (blue, crossed out) are excluded prior to encoding to prevent trivial matching.}
  \label{fig:illustration}
\end{figure*}

\subsection{Unsupervised Adaptation of a Financial Domain Encoder}\label{sec:finetuning}
We aim to train a financial-domain retriever using Form 10-K reports---authoritative and standardized corporate disclosures.
The retriever is designed to retrieve semantically related paragraphs from a large corpus given a query paragraph. 
Unlike re-ranking methods that require pairwise comparisons, our retriever independently encodes each paragraph, enabling scalable and efficient retrieval. 
Each paragraph $p$ is encoded by a transformer-based encoder $f_\theta$, and the final representation $f_\theta(p)$ is obtained by averaging the encoder’s final-layer token embeddings.
Given two paragraphs $p_i$ and $p_j$, we compute their relevance score using the cosine similarity of their vector representations:
\begin{equation}
\label{eq:similarity}
    s(p_i,p_j) = \cos(f_\theta(p_i),f_\theta(p_j)).
\end{equation}

\subsubsection{Unsupervised Training of the Encoder}

\paragraph{Contrastive Learning}
We fine-tune the encoder using contrastive learning to bring similar paragraphs closer in embedding space.
Given an anchor text piece $p$, a positive counterpart $p^+$, and a set of negatives $D^-$, the InfoNCEloss~\cite{oord-etal-2018-cpc} is defined as:
\begin{align*}
    &L(p, p^+, D^-) = \\
    &-\frac{{\exp}^{(s(p,p^+)/\tau)}}{{\exp}^{(s(p,p^+)/\tau)} + \sum\limits_{p^- \in D^-} {\exp}^{s(p,p^-)/\tau}},
\end{align*}
where $\tau$ is a temperature hyperparameter.
Minimizing this loss encourages higher scores to positives and lower scores to negatives.
\label{para:positive-pairs}

\paragraph{Forming Positive Pairs}
We construct high-quality positive pairs using two complementary perspectives: chronological similarity and lexical similarity, both grounded in empirical observations from financial reports.
\begin{itemize}
    \item \textit{Chronological View.} Firms typically experience only one significant event per day (excluding standard accounting dates). 
    We pair paragraphs from the same firm that share identical date-format tokens$[w_{t_1}, w_{t_2}, \dots, w_{t_m}]$ (e.g., ``July 8, 2024''). 
    To avoid superficial matching, we remove all date tokens from both paragraphs during training and validation.
    \item \textit{Lexical View.} Due to regulatory conventions, Form 10-K filings often reuse phrasing to describe similar events.
    We exploit this by creating overlapping spans within the sample paragraph. 
    Given $[w_1, w_2, \dots, w_n]$, we randomly select indices $i<j$ and form the pair $([w_1, \dots, w_j]$, $[w_i, \dots, w_n])$. 
\end{itemize}
An illustrative example for both view is provided in~Figure \ref{fig:illustration}.

\paragraph{Forming Negative Pairs}
Constructing diverse and informative negative pairs is critical for effective contrastive learning. 
We adopt the widely used \emph{in-batch} negatives strategy, as implemented in retrieval models like  Contriever.
This approach treats all other positive samples within the same training batch as negatives, offering a scalable and memory-efficient solution without requiring additional sampling or computation.

Let $D = \{(p_i, p_{i^+})\}_{i=1}^B$ be a batch of  $B$ positive pairs.
For each anchor text piece $p_i$, its positive is $p_{i^+}$, and the set of negatives $D_i^-$ consists of all other positive samples in the batch except  $p_{i^+}$:
\[
D_i^- = \{ p_j \in D \mid j \in \{1^+, 2^+, \dots, B^+\},\ j \ne i^+ \}.
\]
This setup provides $B-1$ negative examples per anchor, enhancing the contrastive signal.
While positive pairs are constructed based on the chronological and lexical similarity views described in Section~\ref{para:positive-pairs}, negative pairs are dynamically formed from the remaining batch entries.
This strategy yields a robust and scalable training procedure, consistent with techniques from prior work~\cite{chen2017sampling, chen2020simclr}.

\subsection{Scoring Mechanism for Risk Relations}\label{sec:scoring}
We utilize our trained encoder to generate paragraph embeddings, where a higher cosine similarity between embeddings indicates potential chronological or lexical alignment.
Given a similarity threshold~$\xi$ (a tuned hyperparameter), we consider two paragraphs $p_i$ and $p_j$ to discuss similar risk content if $s(p_i, p_j) \ge \xi$ (see Eq.~(\ref{eq:similarity})).

Using this similarity criterion, we define the set of \emph{mutual risk paragraphs} (MRPs) between firms $A$ and $B$, denoted as ${\rm MRPs}_{A \leftrightarrow B}$:
\begin{equation*}
\label{eq:mrp-definition}
\begin{aligned}
&{\rm MRPs}_A = \Bigl\{\, p_i \in \mathcal{P}_{A} \,\Big|\,
\exists\, p_j \in \mathcal{P}_{B} : 
s(p_i, p_j) \ge \xi
\Bigr\}, \\
&{\rm MRPs}_B = \Bigl\{\, p_j \in \mathcal{P}_{B} \,\Big|\,
\exists\, p_i\in \mathcal{P}_{A} : 
s(p_i, p_j) \ge \xi
\Bigr\}, \\
&{\rm MRPs}_{A \leftrightarrow B} = {\rm MRPs}_A \cup {\rm MRPs}_B,
\end{aligned}
\end{equation*}
where $\mathcal{P}_{A}$ and $\mathcal{P}_{B}$ are the sets of all paragraphs associated with firms $A$ and $B$, respectively.
MRPs$_A$ (resp.\ MRPs$_B$) consists of paragraphs from firm $A$ (resp. $B$) that share high semantic similarity with at least one paragraph from firm $B$ (resp. $A$). 
These MRPs serve as explicit, interpretable evidence of shared risk exposure.

We then define the \emph{risk relation score} (RRS) between firms $A$ and $B$ as the proportion of mutual risk paragraphs relative to the total number of paragraphs from both firms:
\begin{equation*}
    {\rm RRS}(A,B) = \frac{|{\rm MRPs}_{A \leftrightarrow B}|}{N_A + N_B},
\end{equation*}
where $N_A$ and $N_B$ denote the total numbers of paragraphs from firm $A$ and $B$, respectively. 
The RRS ranges from 0 (no shared risk) to 1 (complete overlap), quantifying the degree of risk-related connection between firms.

\paragraph{Advantages of RRS}
\begin{itemize}
    \item \textbf{Symmetry}: RRS is symmetric by construction, i.e., ${\rm RRS}(A,B) = {\rm RRS}(B,A)$, unlike many retrieval-based methods.
    \item \textbf{Minimum Similarity Guarantee}: By using a threshold-based approach rather than top-$k$ selection, only sufficiently similar paragraph pairs contribute to the score, reducing noise.
    \item \textbf{Interpretability}: Each RRS is grounded by MRPs, allowing transparent inspection of the evidence behind identified risk relations.
\end{itemize}

\section{Experiment}
\label{sec:experiment}
This section presents experimental setup and results.
We begin by detailing our encoder training details, threshold calibration settings, and baseline models. Subsequently, we conduct extensive evaluations designed to answer the following research questions (RQs):
\begin{itemize}
    \item \textbf{RQ1:} How well does the proposed risk relation identification align with real-world stock price co-movements? 
    \item \textbf{RQ2:}What are the individual contributions of the chronological and lexical views to the overall performance?
    \item \textbf{RQ3:} Can the identified risk relations improve downstream tasks such as stock price movement prediction when used as features?
    \item \textbf{RQ4:} How does our dual-view fine-tuned encoder perform on financial information retrieval benchmarks compared to existing models?
\end{itemize}

\subsection{Encoder Training Details}
To prevent information leakage during evaluation, we restrict the training data to Form 10-K filings from 2018 to 2020. 
Filings from all firms and all reported sections of the 10-Ks are included to ensure broad data coverage and diversity.
Each input text piece is truncated or padded to 256 tokens.

For model training, we construct 8,500 positive paragraph pairs and 1,000 validation pairs for each of the two views: chronological and lexical similarity. 
We fine-tune our encoder starting from the BERT-base-uncased pretrained model\footnote{\url{https://huggingface.co/google-bert/bert-base-uncased}}.
Training is performed using contrastive learning with a batch size of 64 and a learning rate of $2\times10^{-5}$, optimized via Adam optimizer~\cite{kingma2014adam} with a linear warmup scheduler.
L2 regularization is applied to improve generalization, and training proceeds for up to 50 epochs with early stopping to mitigate overfitting.\footnote{For computational resources, training was conducted on a single NVIDIA V100 GPU for approximately 8 hours.}

\subsection{RRS Calculation Details }
To ensure reliable identification of shared risks, we apply a threshold-based filtering mechanism that excludes paragraph pairs with insufficient semantic similarity.
The similarity threshold $\xi$ is tuned incrementally in the range [0.5, 0.9] with a step size of 0.05.
The optimal threshold for our encoder is empirically determined to be 0.75, balancing precision and coverage across downstream tasks. We provide additional hyperparameter sensitivity analysis in Appendix~\ref{sec:hyperparameter}.
Moreover, to better focus on risk-related content, we restrict retrieval to paragraphs from Item 1A (Risk Factors) and Item 7A (Quantitative and Qualitative Disclosures about Market Risk) of Form 10-K filings, as these sections are most likely to contain discussions of companies' risk exposures.

\subsection{Encoders for Comparison}\label{sec:encoders}

Since our encoder is fine-tuned from the BERT-base-uncased pretrained model, we include several BERT variants for comparison. 
\begin{itemize}
    \item \textbf{BERT-base-uncased}: The original pretrained model, used as a baseline to measure the effect of our domain-specific fine-tuning.
    \item \textbf{Contriever}~\cite{izacard2021unsupervised}: An unsupervised retrieval model trained with contrastive learning.
    \item \textbf{DPR}~\cite{karpukhin-etal-2020-dpr}: A supervised retrieval model trained on question-answer pairs.
    \item \textbf{FinBERT}~\cite{araci2019finbert} and \textbf{SEC-BERT}~\cite{loukas2022finer}: Two domain-specific models pretrained on financial corpora. We use the base version of SEC-BERT to ensure fairness in model size.
    \item \textbf{Llama-3.2}~\cite{touvron2024llama3}: A widely recognized open-source large language model (LLM). We employ the 3B variant to provide a fair comparison in terms of model size with our encoder.
\end{itemize}
Among these, Llama-3.2 is primarily designed for general-purpose text generation, whereas Contriever and DPR are specifically designed for retrieval tasks. In contrast, FinBERT and SEC-BERT focus on domain adaptation without retrieval-specific objectives.

\label{para:riskrelationevluation}

\begin{table*}
    \centering
    \small
    \setlength{\tabcolsep}{2.8pt}
    \begin{tabular}{lccl c c c c c c}
        \toprule
        & Domain-specific & Retrieval Fine-tuning & Methods & 2020 & 2021 & 2022 & 2023 & 2024 & Avg. \\
        \midrule
        \multirow{2}{*}{Human-based}&&& GICS Sector & 0.1657 & 0.2881 & 0.2964 & 0.2971 & 0.2389 & 0.2572 \\
        &&& GICS Industry & 0.1806 & 0.3336 & 0.2961 & 0.3316 & 0.3115 & 0.2907\\
        \midrule
        \multirow{7}{*}{Model-based}&&& Bert-base-uncased & 0.0804 & 0.2609 & 0.2939 & 0.1945 &  0.2471 & 0.2796 \\
        &&& Llama-3.2-3B & 0.2031 & 0.3965 & 0.3233 & 0.4154 & 0.4336 & 0.3544 \\
        &&\checkmark& Contriever & 0.2136 & 0.3964 & 0.3131 & 0.4112 & 0.4406 & 0.3556 \\
        &&\checkmark& DPR & \underline{0.2138} & \underline{0.4008} & 0.3222 & \underline{0.4158} & \underline{0.4439} & \underline{0.3583}\\
        &\checkmark&& FinBERT & 0.1352 & 0.3013 & 
        0.2706 & 0.2732 & 0.3070 & 0.3058  \\
        &\checkmark&& SEC-BERT & 0.1708 & 0.3545 & \underline{0.3307} & 0.3460 & 0.3633 & 0.3569 \\
        \cmidrule(lr){4-10}
        &\checkmark&\checkmark& Ours & \textbf{0.2141} & \textbf{0.4079} & \textbf{0.3412} & \textbf{0.4233} & \textbf{0.4531} & \textbf{0.3711}\\
        \bottomrule
    \end{tabular}
    \vspace{-0.3cm}
    \caption{{\bf Correlation between RRS and CAVDSR.} {\rm Bold marks the best, underline the second-best.}}
    \label{tab:corr-cavdsr}
\end{table*}
\subsection{Risk Relation Identification Evaluation (RQ1)}
This experiment assesses how well the risk relations identified by our method align with real-world stock price co-movements. 
The underlying intuition is that firms exposed to similar risks often experience correlated stock movements due to the simultaneous impact of common events.
\subsubsection{Data Sources}
We evaluate on firms consistently listed in the S\&P 500 Index from 2018 to 2024, based on the 2024 constituent list.
Dataset comprises stock price data from Yahoo Finance and Form 10-K reports obtained via the SEC API.\footnote{\url{https://www.sec.gov/search-filings/edgar-application-programming-interfaces}}
We retain only firms with complete filings and exclude those involved in mergers or lacking risk disclosures.
After preprocessing to remove all HTML, XBRL tags, and tables, the final dataset covers 2,136 filings from 337 companies.

\subsection{Compared Methods}
We categorize our comparison methods into two groups: Human-based methods and Model-based methods. 
Below, we briefly describe each:
\begin{itemize}
\item \textbf{Human-based Baselines:} The Global Industry Classification Standard (GICS)\footnote{\url{https://www.msci.com/indexes/index-resources/gics}} is a widely adopted taxonomy that assigns each company to exactly one of 11 sectors and 74 industries.
As a human-labeled reference, GICS serves as a proxy for manually defined inter-firm relationships based on industry affiliation.
\item \textbf{Model-based Methods}: In addition to our dual-view fine-tuned encoder, we apply the same risk relation scoring framework to all encoders described in Section~\ref{sec:encoders} to ensure a fair and consistent comparison.
All models follow the same inference procedure: paragraph embeddings are generated by averaging the final-layer token representations, followed by L2 normalization for cosine similarity computation.
\end{itemize}

\subsubsection{Evaluation Metrics} 
\label{sec:eval_metrics}
We propose a metric $\rho$ to measure alignment between risk relation scores (RRSs) and the correlation of the absolute values of daily stock returns (CAVDSR).
Formally:
$\rho = {\rm corr}(\text{RRS}, \text{CAVDSR})$.
CAVDSR is computed using the full year of daily return data for each firm pair, and the corresponding RRS is calculated from the annual 10-K filings of that same pair.
For GICS-based baselines, we assign binary RRS values: 1 if the two firms belong to the same sector or industry, and 0 otherwise.
We use absolute returns to account for divergent effects from the same event (e.g., COVID-19's differing impact on the healthcare sector and the travel sector). 
A higher $\rho$ indicates better alignment with real-world risk co-movement.

\subsubsection{Performance Analysis}
As shown in Table~\ref{tab:corr-cavdsr}, human-based methods yield lower $\rho$, highlighting the limitations of their coarse granularity. Among model-based baselines, retrieval-focused models (DPR and Contriever) generally outperform general-purposed LLM models (Llama-3.2-3b) and other encoders (Bert-base-uncased, FinBERT and SEC-BERT). Specifically, domain-specific models are pretrained on financial text without retrieval-specific objectives, hence underperforming retrieval-focused models.
Our encoder significantly surpasses all baselines, confirming that leveraging chronological and lexical views enhances the identification of shared risk exposures.

\subsection{Ablation Study on Different Views (RQ2)}
To assess the individual contributions of our two training views, we train separate encoders using only the chronological or lexical similarity view.
As shown in Table~\ref{tab:ablation}, the lexical view generally yields stronger performance, suggesting that consistent phrasing in financial reports is particularly effective for capturing risk relationships.

An interesting outcome occurs in 2022, where the chronological-view encoder performs on par with both its lexical-only and dual-view counterparts. 
This result can be attributed to the year's unique market conditions---marked by systemic events such as the Fed's rate hikes, the Russia–Ukraine war, and rising U.S.–China tensions---which triggered widespread, time-aligned impacts across firms.
In such cases, temporal alignment is particularly useful for risk identification.

\begin{table}
\setlength{\tabcolsep}{3pt}
    \small
    \begin{tabular}{l c c c c c}
        \toprule
         & 2020 & 2021 & 2022 & 2023 & 2024 \\
        \midrule
        Chronological & 0.1659 & 0.3691 & \textbf{0.3637} & 0.3671 & 0.3908 \\
        Lexical & \underline{0.2137} & \underline{0.4104} & 0.3336 & \underline{0.4231} & \underline{0.4510}\\
        \midrule
        Ours (both views) & \textbf{0.2161} & \textbf{0.4150} & \underline{0.3421} & \textbf{0.4270} & \textbf{0.4553}\\
        \bottomrule
    \end{tabular}
    \vspace{-0.3cm}
    \caption{{\bf Ablation study on different views.} {\rm Bold indicates the best overall result, while {underline} denotes the second-best.}}
    \label{tab:ablation}
\end{table}

\subsection{Applying Risk Relations to Stock Price Movement Prediction (RQ3)}
We evaluate the practical utility of our risk relation metric by applying it to a downstream financial task: stock price movement prediction.
Specifically, we integrate our risk relations into the attribute-driven graph attention networks (ADGAT)~\cite{cheng-li-2021-adgat} to enhance stock price movement prediction.
The original ADGAT framework notes that using human-defined relations (e.g., sector or industry links) often introduces biases and degrades performance, as such links are static, binary, and lack contextual nuance. 
To address this, we replace ADGAT's predefined relations with those derived from our method. 

\subsubsection{Experimental Setup} 
We closely follow the experimental setup and implementation details of ADGAT.
Specifically, we use 280 days of training data, 70 for validation, and 70 for testing, covering the period from January 1, 2023, to September 4, 2024, with the final 70 days used for evaluation.

For training, we adopt ADGAT's original hyperparameters: Adam optimizer with a learning rate of $5\times 10^{-4}$, a batch size of 15, a dropout rate of 0.2, and training up to 300 epochs, with early stopping.

Each configuration is run 15 times with different seeds, and the averages of the top 5 results are reported.
To evaluate statistical significance, we apply a Mann–Whitney U test~\cite{mann1947u} to compare performance distributions.

\subsubsection{Performance Analysis}
As shown in Table~\ref{tab:downstream}, replacing ADGAT's predefined relations with the ones derived from our model yields a 2.3\% improvement in mean AUC.
This gain is statistically significant ($p < 0.05$) and highlights the value of our method in real-world financial prediction tasks.
\begin{table}
  \centering
  \setlength{\tabcolsep}{13pt}
  \small
  \begin{tabular}{lc}
    \toprule
    \textbf{Method} & \textbf{Mean AUC $\pm$ Std.} \\
    \midrule
    ADGAT (w/o our relation)     & 0.5807 $\pm$ 0.012          \\
    ADGAT (w/ our relation)     & \textbf{0.5939 $\pm$ 0.006}           \\
    \midrule
    Improvement & 2.27\% \\
    \bottomrule
  \end{tabular}
\vspace{-0.3cm}
  \caption{{\bf Performance comparison with/without our relations.} Bold marks the best.}
  \label{tab:downstream}
\end{table}

\begin{table*}
\setlength{\tabcolsep}{5pt}
  \centering
  \small
  \begin{tabular}{lccccccccc}
    \toprule
    \textbf{Model} & \textbf{NDCG@1} & \textbf{NDCG@5} & \textbf{NDCG@10} & \textbf{P@3} & \textbf{P@5} & \textbf{P@10} & \textbf{R@1} & \textbf{R@5} & \textbf{R@10}\\
    \midrule
    Bert-base-uncased & 0.0377 & 0.0159 & 0.0140 & 0.0160 & 0.0103 & 0.0062 & 0.0052 & 0.0079 & 0.0095 \\
    Llama-3.2-3B & 0.0171 & 0.0092 & 0.0089 & 0.0091 & 0.0062 & 0.0038 & 0.0037 & 0.0059 & 0.0070 \\
    Contriever & \underline{0.1712} & \underline{0.0966} & \underline{0.0930} & \underline{0.0902} & \underline{0.0616} & \underline{0.0329} & 0.0394 & \underline{0.0718} & \underline{0.0764} \\
    DPR       & 0.1575 & 0.0819 & 0.0785 & 0.0696 & 0.0480 & 0.0260 & \underline{0.0407} & 0.0581 & 0.0622 \\
    FinBERT     & 0.0034 & 0.0026 & 0.0033 & 0.0023 & 0.0021 & 0.0017 & 0.0009 & 0.0019 & 0.0039 \\
    SEC-BERT  & 0.0274 & 0.0143 & 0.0149 & 0.0137 & 0.0089 & 0.0055 & 0.0071 & 0.0097 & 0.0121 \\
    \midrule
    Ours    & \textbf{0.2021} & \textbf{0.1114} & \textbf{0.1111}& \textbf{0.0993} & \textbf{0.0678} & \textbf{0.0377} & \textbf{0.0518} & \textbf{0.0859} & \textbf{0.0942}\\ 
    \midrule
    Improvement & 18.05\% & 15.32\% & 19.46\% & 10.09\% & 10.09\% & 14.59\% &27.27\% & 19.63\% & 23.30\% \\
    \bottomrule
  \end{tabular}
  \vspace{-0.3cm}
  \caption{\label{tab:MultiHierttResult}
    {\bf Retrieval performance on \texttt{MultiHiertt}.} {\rm Bold marks the best, underline the second-best.}}
\end{table*}

\subsection{Retrieval Performance Evaluation (RQ4)}
Beyond risk relation identification, we evaluate the retrieval effectiveness of our dual-view, financial-domain encoder using \texttt{MultiHiertt} benchmark, comparing it against several strong baselines.

\subsubsection{Data Source}
\texttt{MultiHiertt} is a financial question-answering (QA) benchmark designed to test multi-step numerical reasoning over hierarchical tables.
Although originally proposed for QA, \texttt{MultiHiertt} also serves as a retrieval benchmark, where the task is to retrieve relevant paragraphs from a corpus of 10,475 documents given a query.

\subsubsection{Experimental Setup}
We extract 290 queries and their 1,331 corresponding relevant paragraphs.
A  two-stage retrieval pipeline is adopted.
In the first stage, we retrieve the top 200 documents based on cosine similarities between the query and document embeddings.
In the second stage, the candidates are re-ranked using the BAAI/bge-reranker-v2-m3~\cite{li2023llmretrieval, chen2024bgem3}.
To evaluate our encoder, we replace the retriever component in this pipeline with ours and compare its performance against several baseline encoders.

\subsubsection{Performance Analysis}
Table~\ref{tab:MultiHierttResult} reports standard retrieval metrics such as normalized discounted cumulative gain (NDCG), precision, and recall at various top-$k$ cutoffs.
Our encoder significantly and consistently outperforms all the baselines.
Notably, the two financial-domain encoders, FinBERT and SEC-BERT, perform poorly, likely due to the lack of retrieval-specific fine-tuning.
Similarly, general-domain models (Llama-3.2-3b and BERT-base-uncased) struggle to adapt effectively to financial retrieval tasks.
The key advantage of our encoder lies in its training strategy: the use of dual-view (chronological and lexical) supervision and domain-specific financial data. These design choices enable more effective modeling of semantic relevance in financial retrieval, as reflected in its superior performance.

\section{Case Study}
To demonstrate the practical utility of our method, we present a case study involving Enphase Energy (ENPH) and Meta Platforms (META)---two seemingly unrelated firms from the clean energy and technology sectors, respectively.

In 2023, our model ranks the risk relation between ENPH and META in the 95th percentile, uncovering a shared exposure to supply chain disruptions stemming from the COVID-19 pandemic. 
Although the connection is not immediately obvious, mutual risk paragraphs (MRPs) from their Form 10-K filings reveal a common theme and also provide interpretability for the risk relation between them:
\begin{itemize}
    \item \textbf{Item1A, ENPH:}\textit{ ...The global spread of COVID-19 and other actual or threatened epidemics, pandemics, outbreaks, or public health crises may adversely affect our results of operations and disrupt global supply chains...}
    \item \textbf{Item1A, META:}\textit{ ...We rely on third parties to manufacture and manage the logistics of transporting and distributing our consumer hardware products, which subjects us to a number of risks that have been exacerbated as a result of the COVID-19 pandemic. We have experienced, and may in the future experience, supply or labor shortages or other disruptions in logistics and the supply chain...}
\end{itemize}
External news sources corroborate these findings. 
For instance, the \emph{Financial Times} reported Meta’s hardware delays due to supply chain shocks,\footnote{\url{https://www.ft.com/content/c7e9cfa9-3f68-47d3-92fc-7cf85bcb73b3}} 
while \emph{pv magazine USA} highlighted persistent pandemic-related disruptions affecting the solar industry.\footnote{\url{https://pv-magazine-usa.com/2023/01/04/three-solar-industry-trends-to-watch-in-2023/}}
This example highlights the encoder's ability to uncover subtle, non-obvious risk links with real-world relevance. A higher risk relation score (RRS) indicates that two companies are closely connected through shared risk exposures (e.g. supply chain disruptions), as demonstrated in the case above. These risk relations provide valuable interpretability, offering meaningful insights that can support more informed financial decision-making.
\section{Conclusion} \label{sec:conclusion}
This paper proposes a novel framework for identifying inter-firm risk relations directly from unstructured financial text.
By leveraging chronological and lexical similarities in Form 10-K filings, we develop an unsupervised fine-tuning strategy and introduce a transparent, symmetric risk relation core (RRS) to quantify shared exposures.

Extensive experiments validate the effectiveness of our method: (1) RRS shows strong correlation with real-world stock price co-movements, (2) its integration improves downstream stock prediction in a graph-based model (ADGAT), and (3) the encoder achieves superior performance on the \texttt{MultiHiertt} retrieval benchmark.
A case study further demonstrates the method’s ability to uncover subtle but meaningful risk connections.

\section{Limitations}
Although our method demonstrates strong performance in identifying nuanced inter-firm risk relationships, several limitations should be acknowledged.
First, the framework is designed specifically to uncover relations based on shared risk exposures.
As such, it may not generalize well to tasks
involving other types of firm interactions, such as strategic partnerships, mergers and acquisitions, or product-market complementarities.

Second, our current approach relies solely on Form 10-K filings as the data source.
Although these documents are structured and reliable, their annual frequency limits our method's responsiveness to short-term market changes or evolving risk profiles within a fiscal year.
This restricts its applicability for real-time or high-frequency financial decision-making.

Lastly, while our evaluation leverages public financial and market data, it does not incorporate expert financial judgment. Practical decision-making often involves qualitative insights and domain expertise that automated models alone cannot fully capture. Future work could benefit from integrating expert input to enhance interpretability and real-world applicability.

\bibliography{custom}

\appendix
\begin{figure}[t]
  \centering
  \includegraphics[width=0.45\textwidth]{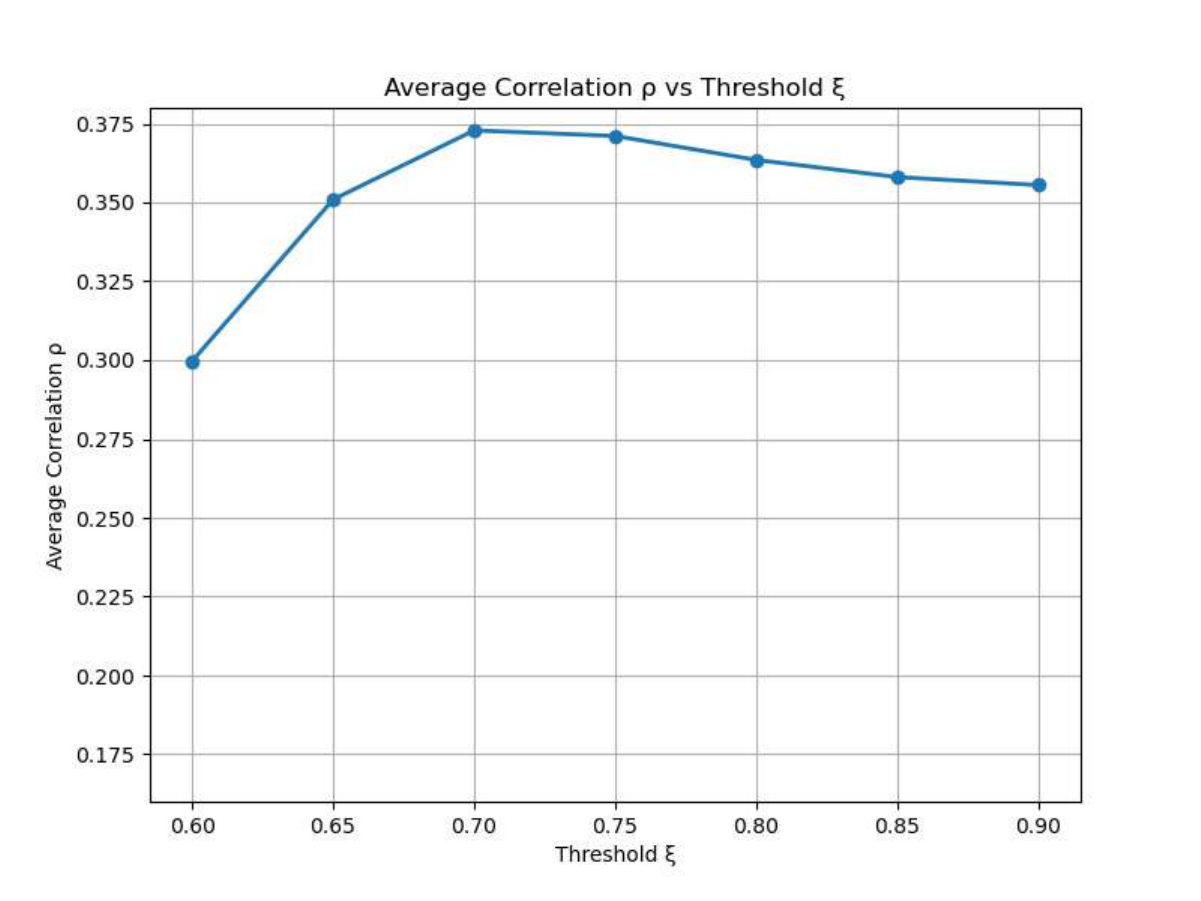}
  \caption{\bf{Hyperparameter Sensitivity Analysis}}
  \label{fig:sensitivity}
\end{figure}
\section{Hyperparameter Sensitivity Analysis}
\label{sec:hyperparameter}
We conduct a sensitivity analysis on the similarity threshold used by our encoder for identifying mutual risk paragraphs. Specifically, the threshold is tuned from 0.6 to 0.9, and the results of the first experiment described in Section~\ref{sec:eval_metrics} are reported in Figure~\ref{fig:sensitivity}. The results show that our encoder’s performance remains relatively stable across different threshold settings, highlighting both its robustness and effectiveness in consistently identifying inter-firm risk relations.


\end{document}